\pdfoutput=1
\documentclass[conference, 10pt]{IEEEtran}
\IEEEoverridecommandlockouts

\usepackage{color}
\usepackage{theorem}
\usepackage{times,amsmath,epsfig}
\usepackage{amssymb}
\usepackage{subcaption}
\usepackage{cite}
\usepackage{url}
\usepackage[capitalise]{cleveref}
\usepackage[ruled,linesnumbered]{algorithm2e}
\usepackage{tikz, pgfplots}
\usepackage{moresize}
\usepackage{array}

\input{mysymbol.sty}

\pgfplotsset{compat=1.17}
\pgfplotstableset{col sep=semicolon}
\usepgfplotslibrary{fillbetween}
\tikzset{every mark/.append style={scale=1.5, solid}, font=\small}
\pgfplotsset{
    width=1.05\textwidth,
    height=5.5cm,
    legend style={
        font=\ssmall ,  
        inner xsep=1pt,
        inner ysep=1pt,
        nodes={inner sep=1pt}},
    legend cell align=left,
    every axis/.append style={line width=.5pt},
 	every axis plot/.append style={line width=1.5pt},
 	every axis y label/.append style={yshift=-4pt}
}

\title{{Convolutional Learning on Directed Acyclic Graphs}}
\author{{Samuel Rey$^*$,
        Hamed Ajorlou$^\dagger$
        and~Gonzalo Mateos$^\dagger$} \\
        $^*$Dept. of Signal Theory and Communications,
        Rey Juan Carlos University, Madrid, Spain\\
        Email: samuel.rey.escudero@urjc.es. \\
        $^\dagger$Dept. of Electrical and Computer Engineering, University of Rochester, NY, USA. \\
        Emails: \{gmateosb, hajorlou\}@ur.rochester.edu
        
\thanks{{This work was supported in part by
the Spanish AEI under Grants PID2019-105032GBI00, TED2021-130347B-I00 and PID2022-136887NBI00 funded by MCIN/AEI/10.13039/501100011033.}}}

\begin{document}
%
\maketitle
\begin{abstract}
We develop a novel convolutional architecture tailored for learning from data defined over directed acyclic graphs (DAGs). DAGs can be used to model causal relationships among variables, but their nilpotent adjacency matrices pose unique challenges towards developing DAG signal processing and machine learning tools. To address this limitation, we harness recent advances offering alternative definitions of causal shifts and convolutions for signals on DAGs. We develop a novel convolutional graph neural network that integrates learnable DAG filters to account for the partial ordering induced by the graph topology, thus providing valuable inductive bias to learn effective representations of DAG-supported data. We discuss the salient advantages and potential limitations of the proposed DAG convolutional network (DCN) and evaluate its performance on two learning tasks using synthetic data: network diffusion estimation and source identification. DCN compares favorably relative to several baselines, showcasing its promising potential.
\end{abstract}
\begin{IEEEkeywords}
DAG, graph signal processing, graph neural networks, graph convolution.
\end{IEEEkeywords}

\section{Introduction}\label{s:introduction}
The vast complexity and heterogeneity of modern networks are propelling the generation of data represented on non-Euclidean domains.
Consequently, fields such as graph signal processing (GSP) have emerged, which model the underlying irregular signal structure as a graph and then exploit its topology to process said relational data~\cite{ortega2018graph,dong2020graph}.
Graph neural networks (GNN) are non-linear, graph-aware machine learning (ML) models that serve as a prominent example of this new paradigm~\cite{bronstein2017geometric,rey2019underparametrized,wu2020comprehensive,tenorio2021robust}. 
GNNs exploit prior information in the graph via convolutions~\cite{kipf2016semi,xu2018powerful,ruiz2021graph}, attention mechanisms~\cite{velivckovic2018graph}, or graph autoencoders~\cite{wang2017mgae,rey2021overparametrized}, and have attained state-of-the-art performance in a gamut of ML applications.

Arguably most GNNs and graph-based methods focus on undirected graphs. However, accounting for directionality can play an important role in processing information, but this leap comes with well-documented challenges~\cite{shafipour2018dgft,shafipour2018directed,marques2020signal} that are exacerbated when dealing with a directed acyclic graph (DAG).
For instance, DAGs have nilpotent adjacency matrices and hence a collapsed spectrum that prevents us from directly applying spectral-based tools~\cite{seifert2023causal}.  
These difficulties notwithstanding, DAGs are prevalent across domains including causal inference~\cite{peters2017elements}, learning Bayesian networks and structural causal models~\cite{peters2014identifiability,zheng2018dags,dacunto2023multiscale,saboksayr2023colide}, code parsing~\cite{allamanis2018survey}, or performance prediction for neural architectures~\cite{zhang2018graph}, just to name a few.

This work develops a novel \emph{convolutional} GNN tailored for learning from signals defined over DAGs.
As DAGs impose a partial ordering on the set of nodes, a key challenge is to incorporate this stronger inductive bias into our architecture in a principled manner.
To that end, we build on a recent framework that extends GSP tools to partially ordered sets (posets)~\cite{puschel2021discrete,seifert2023causal}, to arrive at a DAG convolutional network (DCN). These foundational design principles can be traced to linear structural equation models (SEMs), making DCN well-suited to capture causal relations in the data. 
Moreover, the resulting architecture admits a spectral representation and is permutation equivariant.
Unlike related alternatives~\cite{zhang2019advances,thost2021directed}, the convolutional layers primarily involve sparse matrix multiplications, resulting in manageable computational complexity.

\noindent \textbf{Related work and contributions in context.} Developing GNNs to process data defined over DAGs is an important problem that is starting to draw attention.
Recently,~\cite{zhang2019advances} advocated embedding DAGs into a smooth latent space and introduced a variational autoencoder for DAGs, dubbed D-VAE.
The DAG neural network (DAGNN) from~\cite{thost2021directed} sequentially aggregates the node representations from predecessors via an attention mechanism, and then uses a gated recurrent unit (GRU) to encode the causal structure of the DAG during a combination step.
However, the sequential scheme followed by both D-VAE and DAGNN results in a large computational burden, limiting their applicability to moderately-sized graphs.
Furthermore, DAGNN needs to combine the true and reverse directions of the DAG in a non-intuitive way to process the data.
Later,~\cite{luo2023transformers} proposed encoding the DAG ordering via a reachability-based attention mechanism, extending graph transformers to DAGs.

All in all, the main contributions of this work are as follows:
\begin{itemize}
    \item We propose DCN, the first convolutional GNN designed to learn from data defined over DAGs. We elaborate on the key benefits of the model and discuss its limitations. 
    \item Different from existing approaches~\cite{zhang2019advances,thost2021directed}, DCN layers rely on formal convolutions derived in~\cite{seifert2023causal} for a signal model over posets, that can be linked to linear SEMs.
    \item We evaluate the performance of DCN and show it compares favorably to several baselines in different graph learning tasks. We share the code to reproduce our results. 
\end{itemize}

\section{Preliminaries and problem statement}\label{s:fundamentals}
We first define basic concepts and notation about DAGs, as well as the required background on GSP and GNNs. We then formally state the problem of learning from DAG signals. 

\subsection{Preliminaries: DAGs, graph signals and GNNs}

\noindent \textbf{DAGs and graph signals.}
Let $\ccalD = (\ccalV, \ccalE)$ be a directed acyclic graph (DAG), where $\ccalV$ denotes the set of $N$ nodes and $\ccalE \subseteq \ccalV \times \ccalV$ denotes the set of edges.
An edge $(i,j) \in \ccalE$ exists if and only if there is a link from node $j$ to node $i$.
Since $\ccalD$ contains no cycles, we can sort $\ccalV$ topologically, ensuring that node $j$ precedes node $i$ whenever $(i,j) \in \ccalE$.
Consequently, every DAG induces a unique partial order on $\ccalV$, where $j < i$ if $j$ is a \emph{predecessor} of $i$, i.e., if there exists a path from $j$ to $i$~\cite{puschel2021discrete}.
Note that in posets not all the elements are comparable.
In fact, for every pair $i, j \in \ccalV$ without a path from $i$ to $j$ or vice-versa, we have that $i \not \leq j$ and $j \not \leq i$.

Let $\bbA \in \reals^{N\times N}$ denote the (possibly weighted) adjacency matrix of $\ccalD$, where $A_{ij} \neq 0$ if and only if $(i,j) \in \ccalE$.
Due to the aforementioned partial ordering of $\ccalV$, we can sort the entries of $\bbA$ to obtain a strictly lower-triangular matrix.
The diagonal is null because DAGs cannot have self-loops.

In addition to the graph $\ccalD$, we consider signals defined on the set of nodes $\ccalV$.
Formally, graph signals are functions from the vertex set to the real field $x: \ccalV \mapsto \reals$, which can be represented as a vector $\bbx \in \reals^N$, where $x_i$ is the signal value at node $i$. A signal processing theory can be constructed for graph signals supported on DAGs (a.k.a. DAG signals), which was put forth in~\cite{seifert2023causal} and we review in Section \ref{ssec:DAGSP}.

\vspace{2mm}
\noindent \textbf{Convolutional GNNs.}
A GNN can be represented as a parametric non-linear function that depends on the underlying graph structure.
While several GNN models have been proposed, most of them build upon an \emph{aggregation function} driven by the graph topology.
In this work we consider a class of GNNs where the aggregation is accomplished through graph convolutions~\cite{kipf2016semi,ruiz2021graph}.
More precisely, we are interested in GNNs employing a bank of learnable convolutional graph filters, where the output of layer 
$\ell=1,\ldots,L$ is given by
\begin{equation}\label{eq:fb_gnn}
    \bbX^{(\ell+1)} = \sigma \left( \sum_{r=0}^{R-1} \bbA^r \bbX^{(\ell)} \bbTheta_r^{(\ell)} \right).
\end{equation}
Here, $\bbTheta_r^{(\ell)} \in \reals^{F_i^{(\ell)} \times F_o^{(\ell)}}$ collects the learnable filter coefficients, $F_i^{(\ell)}$ and $F_o^{(\ell)}$ are the number of input and output features of the $\ell$-th layer, and the powers $\bbA^r$ determine the radius of the aggregation neighborhood.
If necessary, the adjacency matrix could be substituted with any desired \emph{graph-shift operator} (GSO) that encodes the graph structure; see e.g.,~\cite{ortega2018graph,ruiz2021graph}. The point-wise nonlinear activation function is often chosen to be a ReLU, i.e., $\sigma(x)=\max(0,x)$.

\subsection{Problem statement}
This work addresses the problem of learning from signals defined over a DAG $\ccalD$.
We are given $\ccalT = \{ \bbX_m, \bby_m \}_{m=1}^M$, the training set containing $M$ input-output observations from a network process on $\ccalD$.
The input graph signals are denoted as $\bbX_m \in \reals^{N \times F}$ and the corresponding outputs as $\bby_m \in \reals^N$.
Here outputs are single-feature graph signals $\bby_m$ for simplicity, and generalizations to multi-feature signals or nodal/graph labels are straightforward.
We use $\ccalT$ to learn the map relating $\bbX_m$ and $\bby_m$, which is assumed to be accurately represented by a non-linear parametric function $f_{\bbTheta}(\cdot | \ccalD): \reals^{N\times F}\mapsto \reals^N$.
To that end, we estimate the weights $\bbTheta$ of $f_{\bbTheta}(\cdot | \ccalD)$ by solving
\begin{equation}
    \min_{\bbTheta} \frac{1}{M}\sum_{m=1}^M \ccalL(\bby_m, f_{\bbTheta} (\bbX_m | \ccalD)),
\end{equation}
%
where $\ccalL$ is a loss function, e.g., mean squared error (MSE) for regression or the cross entropy loss for classification.

\section{Convolutional learning on DAGs}
Among the many graph-aware functions available, we confine our selection of $f_{\bbTheta}(\cdot | \ccalD)$ to the class of graph convolutional neural networks (GCNN); see e.g.,~\cite{kipf2016semi,ruiz2021graph} for extensive discussion and justification on their merits.
The first step towards designing a GCNN tailored to DAG signals is to define the convolution operation.
Naively, one could use the graph filters in \eqref{eq:fb_gnn}, as it is customary when graphs have cycles~\cite{segarra2017optimal,ruiz2021graph}.
However, for a DAG all the eigenvalues of $\bbA$ are 0, due to its strict lower triangular structure.
This collapsed spectrum deprives us of a spectral interpretation for such filters.
Moreover, recall that DAGs impose a partial ordering and can potentially capture causal relations, two properties we wish to encode in our GCNN.
To overcome these challenges and design $f_{\bbTheta}(\cdot | \ccalD)$ in a principled manner (Section \ref{ssec:DCN}), we bring to bear the novel DAG signal processing framework introduced in~\cite{seifert2023causal}, which we briefly outline next.

\subsection{Graph-shift operators and convolution for DAGs}\label{ssec:DAGSP}
Let $\bbc \in \reals^N$ be a vector collecting the causes or nodal contributions of a DAG signal $\bbx$.
The signal model introduced in~\cite{seifert2023causal} postulates that $\bbx$ can be described by the causes at predecessor nodes. Specifically, we write $\bbx = \bbW \bbc$,
where $\bbW \in \reals^{N \times N}$ is the \emph{weighted transitive closure} of $\ccalD$ with $W_{ij} \neq 0$ if there is a path from $j$ to $i$ (i.e., if $j < i$) or $i = j$.
Note that $\bbW$ can be interpreted as the adjacency matrix of the reachability graph of $\ccalD$, with ones on the diagonal to account for the reflexivity property of posets.
Among the different methods to compute $\bbW$ from $\bbA$~\cite{lehmann1977algebraic}, we focus on $\bbW = (\bbI - \bbA)^{-1}$.
This is equivalent to a linear SEM for $\bbx$, with exogenous inputs $\bbc$~\cite[Section IV-D]{seifert2023causal}.

Given this model, every node $k \in \ccalV$ induces a causal GSO on the signal $\bbx$ given by the matrix $\bbT_k \in \reals^{N \times N}$, such that 
\begin{equation}\label{eq:gso}
    [\bbT_k\bbx]_i = \sum_{j \leq i \; \mathrm{and} \; j \leq k} W_{ij}c_j.
\end{equation}
Intuitively, the signal value at node $i$ after shifting $\bbx$ with respect to node $k$ results in a signal containing only causes from common predecessors to both $i$ and $k$.
This can be compactly written in matrix form as
\begin{equation}\label{eq:compact_gso}
    \bbT_k\bbx = \bbW \bbD_k \bbc = \bbW \bbD_k \bbW^{-1} \bbx,    
\end{equation}
where $\bbD_k$ is a diagonal matrix whose entry $[\bbD_k]_{ii}=1$ if $i \leq k$ (i.e., if node $i$ is a predecessor of node $k$) and 0 otherwise.
The inverse $\bbW^{-1}$ always exists since $\bbW$ is a full-rank matrix and it can be efficiently computed using the weighted Moebius inversion~\cite{rota1964foundations}.
It follows that the columns of $\bbW$ and the diagonal entries of $\bbD_k$ are, respectively, the eigenvectors and eigenvalues of the GSO $\bbT_k$. Matrix $\bbW$ is thus a Fourier basis for DAG signals, while $\bbW^{-1}$ is the corresponding Fourier transform. Because $\bbx=\bbW\bbc$, for this DAG signal model the causes $\bbc$ are also spectral coefficients.

In GSP, a graph filter is a polynomial of the (single) GSO.
In contrast, now we have a GSO $\bbT_k$ for every node in $\ccalV$ and, moreover, every $\bbT_k$ is an idempotent operator. 
Consequently, the most general shift-invariant DAG filter $\bbH$ is given by
\begin{equation}\label{eq:dag_filter}
    \bbH = \sum_{k \in \ccalV} h_k \bbT_k = \bbW \sum_{k \in \ccalV} h_k \bbD_k \bbW^{-1},
\end{equation}
%
resulting in a convolution operation $\bbh*\bbx = \bbH\bbx$.
Here, $\bbh\in \reals^N$ collects the filter coefficients $h_k$, and the frequency response of $\bbH$ is given by the diagonal of $\sum_{k \in \ccalV} h_k \bbD_k$.
This definition of DAG filter plays a central role in our convolutional architecture, the subject dealt with next.

\subsection{DAG Convolutional Network (DCN)}\label{ssec:DCN}
Now, we are ready to introduce our DAG-aware convolutional network, which we refer to as DCN.
The proposed DCN harnesses the definition of causal graph filters in \eqref{eq:dag_filter} to implement convolutions, which we compose with a point-wise nonlinearity to process DAG signals. In the simplest case, the output of layer 
$\ell=1,\ldots,L$ can be computed as 
\begin{equation}\label{eq:dcn_perceptron}
    \bbx^{(\ell+1)} = \sigma \left( \sum_{k \in \ccalV} h_k^{(\ell)} \bbT_k \bbx^{(\ell)}\right),
\end{equation}
where $h_k^{(\ell)}$ are the learnable filter coefficients.

We refer to \eqref{eq:dcn_perceptron} as a DAG perceptron, and we can gain intuition from two different standpoints.
In the spectral domain, recall that $\bbT_k\bbx^{(\ell)} = \bbW\bbD_k\bbc^{(\ell)}$.
In words, the $k$-th GSO selects the causes (i.e., the exogenous input in an SEM) of predecessors of the node $k$, and diffuses them across the reachability graph via $\bbW$.
Clearly, this accounts for the partial ordering imposed by the DAG.
Alternatively, one could look at \eqref{eq:dcn_perceptron} in the context of message-passing networks.
For every node $i \in \ccalV$, each $\bbT_k$ forms a message combining the features from predecessors common to $i$ and $k$, to finally yield $x_i^{(\ell+1)}$ as a non-linear weighted combination of these messages.
A complete example and schematic illustrating these layer operations will be included in the extended journal version.

While \eqref{eq:dcn_perceptron} is intuitive, a single filter may not be expressive enough to learn complex DAG signal representations.
Furthermore, input signals typically comprise several features.
We address these limitations following a rationale similar to that of \eqref{eq:fb_gnn}. At each layer, the single filter is replaced with a bank of learnable DAG filters, resulting in the recursion [cf. \eqref{eq:fb_gnn}]
\begin{equation}\label{eq:fb_dagnn}
    \bbX^{(\ell+1)} = \sigma \left( \sum_{k \in \ccalV} \bbT_k \bbX^{(\ell)} \bbTheta_k^{(\ell)}\right).
\end{equation}
Here, the matrices $\bbTheta_k^{(\ell)} \in \reals^{F_i \times F_o}$ collect the learnable filter coefficients with $F_i$ and $F_o$ denoting the number of input and output features.
Then, for a DCN with $L$ layers, the prediction $\hby_m = \bbX^{(L)} = f_{\bbTheta} (\bbX_m | \ccalD)$ will be the output of the final layer.

\begin{remark}
    The DCN whose output is given by the recursion \eqref{eq:fb_dagnn} is permutation equivariant.
    Due to space limitations, the proof is deferred to the extended journal version of this work.
\end{remark}

\noindent \textbf{Discussion.} The DCN convolutional layer \eqref{eq:fb_dagnn} is closely related to GCNN architectures for general (cyclic) graphs.
However, relying on causal graph filters is imperative to implement formal convolutions, and offers some key advantages.
First, the spectrum of $\bbT_k$ is well defined, endowing the DCN with a spectral representation that is fundamental to analyze properties such as stability, transferability, or denoising capability~\cite{ruiz2021graph, rey2022untrained}, an interesting future research direction.
Furthermore, since the eigenvalues of $\bbT_k$ are the binary matrices $\bbD_k$, no issues stemming from numerical instability are expected when concatenating several layers.

On the flip side, the number of GSOs potentially involved in the convolution -- hence the number of learnable parameters -- grows with the size of the graph. This may lead to computational and memory limitations.
As an engineering workaround, one can approximate and simplify the convolution in \eqref{eq:fb_dagnn} as $\sum_{k \in \ccalU} h_k \bbT_k$, where $\ccalU \subset \ccalV$.
In principle, this comes at the expense of reducing the expressiveness of the DCN. Still, this limitation can be alleviated by concatenating several layers since the cross-product of different $\bbT_k$ and $\bbT_l$ can give rise to a new GSO.
As we illustrate in the upcoming numerical evaluation, in practice one can randomly select a small number of matrices $\bbT_k$ to define the convolution and maintain a competitive performance.
Developing a principled approach to determine the subset of nodes $\ccalU$ constitutes an interesting problem that is considered a future research direction.

\begin{table}[]
\centering
\begin{tabular}{l|c c | c c}
 & \multicolumn{2}{c|}{\textbf{Network Diffusion}} & \multicolumn{2}{c}{\textbf{Source Identification}} \\
 & MNSE & Time (s) & Accuracy & Time (s) \\
\hline
DCN & $\mathbf{0.016 \pm 0.014}$ & 3.6 & \(0.052 \pm 0.014\) & 7.5 \\
DCN-30 & $\mathbf{0.029 \pm 0.017}$ & 3.5 & \(0.052 \pm 0.016\) & 7.4 \\
DCN-10 & \(0.058 \pm 0.021\) & 3.5 & \(0.055 \pm 0.015\) & 7.2 \\
\hline
DCN-T & \(0.098 \pm 0.024\) & 4.1 & $\mathbf{0.991 \pm 0.018}$ & 8.2 \\
DCN-30-T & \(0.199 \pm 0.030\) & 3.7 & $\mathbf{0.983 \pm 0.032}$ & 7.64 \\
DCN-10-T & \(0.229 \pm 0.030\) & 3.5 & \(0.865 \pm 0.141\) & 7.38 \\
\hline
LS & \(0.050 \pm 0.022\) & 0.4 & \(0.05 \pm 0.016\) & 0.36 \\
FB-GCNN & \(0.091 \pm 0.028\) & 3.4 & \(0.739 \pm 0.172\) & 7.4 \\
GCN & \(0.167 \pm 0.037\) & 3.3 & \(0.155 \pm 0.216\) & 7.1 \\
GAT & \(0.649 \pm 0.089\) & 13.8 & \(0.044 \pm 0.081\) & 28.4 \\
GraphSAGE & \(0.359 \pm 0.039\) & 5.9 & \(0.676 \pm 0.163\) & 12.5 \\
GIN & \(0.402 \pm 0.079\) & 6.0 & \(0.19 \pm 0.163\) & 12.5 \\
MLP & \(0.353 \pm 0.039\) & 2.2 & \(0.050 \pm 0.016\) & 4.7 \\
\hline
\end{tabular}
\caption{Normalized MSE and accuracy when respectively solving the network diffusion learning and source identification tasks.
We report the mean performance and standard deviation across 25 realizations, along with the training time.}
\vspace{-5mm}
\label{tab:summary_results}
\end{table}


\section{Numerical Evaluation}\label{s:numerical_eval}
We assess the performance of DCN in different settings and compare it with several baselines.
Code to reproduce all results is available on GitHub\footnote{\url{https://github.com/reysam93/dag_conv_nn}}, and the interested reader is referred there for additional experiments and implementation details.

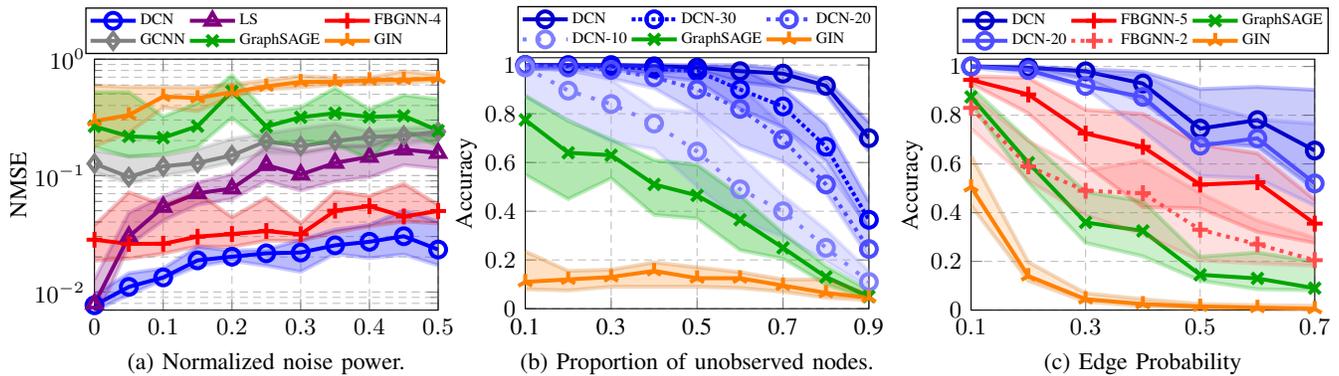
\begin{figure*}[!t]
	\centering
	\begin{subfigure}{0.32\textwidth}
		\centering
		 \begin{tikzpicture}[baseline,scale=1]

\pgfplotstableread{data/noise/noise_inf-constant-med_err.csv}\errtable
\pgfplotstableread{data/noise/noise_inf-constant-prct75_err.csv}\prcttop
\pgfplotstableread{data/noise/noise_inf-constant-prct25_err.csv}\prctbot

\pgfmathsetmacro{\opacity}{0.3}
\pgfmathsetmacro{\contourop}{0.25}

\begin{semilogyaxis}[
    xlabel={(a) Normalized noise power.},
    xmin=0,
    xmax=0.5,
    xtick = {0, .1, ..., .51},
    ylabel={NMSE},
    ymin = .007,
    ymax = 1,
    grid style=densely dashed,
    grid=both,
    legend style={
        at={(.5, 1.02)},
        anchor=south},
    legend columns=3,
    width=175,
    height=140,
    ]

    \addplot [blue!80!white, name path = DCN-bot, opacity=\contourop, forget plot] table [x=xaxis, y=DCN] \prctbot;
    \addplot [blue!90!white, name path = DCN-top, opacity=\contourop, forget plot] table [x=xaxis, y=DCN] \prcttop;
    \addplot[blue!90!white, fill opacity=\opacity, forget plot] fill between[of=DCN-bot and DCN-top];
    \addplot[blue, mark=o, solid] table [x=xaxis, y=DCN] {\errtable};

    \addplot [violet!90!white,name path = Linear-bot, opacity=\contourop, forget plot] table [x=xaxis, y=Linear] \prctbot;
    \addplot [violet!90!white,name path = Linear-top, opacity=\contourop, forget plot] table [x=xaxis, y=Linear] \prcttop;
    \addplot[violet!90!white, fill opacity=\opacity, forget plot, forget plot] fill between[of=Linear-bot and Linear-top];
    \addplot[violet, mark=triangle , solid] table [x=xaxis, y=Linear] {\errtable};
    
    \addplot [red!80!black, name path = FB-GCNN-bot, opacity=\contourop, forget plot] table [x=xaxis, y=FB-GCNN-4] \prctbot;
    \addplot [red!80!black, name path = FB-GCNN-top, opacity=\contourop, forget plot] table [x=xaxis, y=FB-GCNN-4] \prcttop;
    \addplot[red!80!white, fill opacity=\opacity, forget plot, forget plot] fill between[of=FB-GCNN-bot and FB-GCNN-top];
    \addplot[red, mark=+, solid] table [x=xaxis, y=FB-GCNN-4] {\errtable};

    \addplot [gray!90!white,name path = GNN-bot, opacity=\contourop, forget plot] table [x=xaxis, y=GNN-A] \prctbot;
    \addplot [gray!90!white,name path = GNN-top, opacity=\contourop, forget plot] table [x=xaxis, y=GNN-A] \prcttop;
    \addplot[gray!90!white, fill opacity=\opacity, forget plot, forget plot] fill between[of=GNN-bot and GNN-top];
    \addplot[gray, mark=diamond , solid] table [x=xaxis, y=GNN-A] {\errtable};

    \addplot [green!70!black,name path = GraphSAGE-bot, opacity=\contourop, forget plot] table [x=xaxis, y=GraphSAGE-A] \prctbot;
    \addplot [green!70!black,name path = GraphSAGE-top, opacity=\contourop, forget plot] table [x=xaxis, y=GraphSAGE-A] \prcttop;
    \addplot[green!70!black, fill opacity=\opacity, forget plot, forget plot] fill between[of=GraphSAGE-bot and GraphSAGE-top];
    \addplot[green!65!black, mark=x, solid] table [x=xaxis, y=GraphSAGE-A] {\errtable};
    
    \addplot [orange!80!black, name path = GIN-bot, opacity=\contourop, forget plot] table [x=xaxis, y=GIN-A] \prctbot;
    \addplot [orange!80!black, name path = GIN-top, opacity=\contourop, forget plot] table [x=xaxis, y=GIN-A] \prcttop;
    \addplot[orange!80!white, fill opacity=\opacity, forget plot, forget plot] fill between[of=GIN-bot and GIN-top];
    \addplot[orange, mark=Mercedes star, solid] table [x=xaxis, y=GIN-A] {\errtable};

    \legend{DCN, LS, FBGNN-4, GCNN, GraphSAGE, GIN}
\end{semilogyaxis}
\end{tikzpicture}\label{fig:experiment_a}
	\end{subfigure}
	\begin{subfigure}{0.32\textwidth}
		\centering
        \begin{tikzpicture}[baseline,scale=1]

\pgfplotstableread{data/src_nodes/src_nodes_inf-random-med_acc.csv}\acctable
\pgfplotstableread{data/src_nodes/src_nodes_inf-random-prct75_acc.csv}\prcttop
\pgfplotstableread{data/src_nodes/src_nodes_inf-random-prct25_acc.csv}\prctbot

\pgfmathsetmacro{\opacity}{0.3}
\pgfmathsetmacro{\contourop}{0.25}

\begin{axis}[
    xlabel={(b) Proportion of unobserved nodes.},
    xmin=10,
    xmax=90,
    xtick = {10, 30, ..., 90},
    xticklabels = {0.1, 0.3, 0.5, 0.7, 0.9},
    ylabel={Accuracy},
    ymin = 0,
    ymax = 1.03,
    ytick = {0, .2, ..., 1},
    grid style=densely dashed,
    grid=both,
    legend style={
        at={(.5, 1.02)},
        anchor=south},
    legend columns=3,
    width=175,
    height=140,
    ]

    \addplot [blue!90!black, name path = DCN-bot, opacity=\contourop, forget plot] table [x=xaxis, y=DCN] \prctbot;
    \addplot [blue!90!black, name path = DCN-top, opacity=\contourop, forget plot] table [x=xaxis, y=DCN] \prcttop;
    \addplot[blue!90!black, fill opacity=\opacity, forget plot] fill between[of=DCN-bot and DCN-top];
    \addplot[blue!80!black, mark=o, solid] table [x=xaxis, y=DCN] {\acctable};
    
    \addplot [blue!90!white,name path = DCN-30-bot, opacity=\contourop, forget plot] table [x=xaxis, y=DCN-30] \prctbot;
    \addplot [blue!90!white,name path = DCN-30-top, opacity=\contourop, forget plot] table [x=xaxis, y=DCN-30] \prcttop;
    \addplot[blue!90!white, fill opacity=\opacity, forget plot] fill between[of=DCN-30-bot and DCN-30-top];
    \addplot[blue, mark=o, densely dotted] table [x=xaxis, y=DCN-30] {\acctable};

    \addplot [blue!60!white, name path = DCN-20-bot, opacity=\contourop, forget plot] table [x=xaxis, y=DCN-20] \prctbot;
    \addplot [blue!60!white, name path = DCN-20-top, opacity=\contourop, forget plot] table [x=xaxis, y=DCN-20] \prcttop;
    \addplot[blue!60!white, fill opacity=\opacity, forget plot] fill between[of=DCN-20-bot and DCN-20-top];
    \addplot[blue!70!white, mark=o, dotted] table [x=xaxis, y=DCN-20] {\acctable};

    \addplot [blue!40!white, name path = DCN-10-bot, opacity=\contourop, forget plot] table [x=xaxis, y=DCN-10] \prctbot;
    \addplot [blue!40!white, name path = DCN-10-top, opacity=\contourop, forget plot] table [x=xaxis, y=DCN-10] \prcttop;
    \addplot[blue!40!white, fill opacity=\opacity, forget plot, forget plot] fill between[of=DCN-10-bot and DCN-10-top];
    \addplot[blue!50!white, mark=o, loosely dotted] table [x=xaxis, y=DCN-10] {\acctable};

    \addplot [green!70!black,name path = GraphSAGE-bot, opacity=\contourop, forget plot] table [x=xaxis, y=GraphSAGE-A] \prctbot;
    \addplot [green!70!black,name path = GraphSAGE-top, opacity=\contourop, forget plot] table [x=xaxis, y=GraphSAGE-A] \prcttop;
    \addplot[green!70!black, fill opacity=\opacity, forget plot, forget plot] fill between[of=GraphSAGE-bot and GraphSAGE-top];
    \addplot[green!65!black, mark=x, solid] table [x=xaxis, y=GraphSAGE-A] {\acctable};

    
    \addplot [orange!80!black, name path = GIN-bot, opacity=\contourop, forget plot] table [x=xaxis, y=GIN-A] \prctbot;
    \addplot [orange!80!black, name path = GIN-top, opacity=\contourop, forget plot] table [x=xaxis, y=GIN-A] \prcttop;
    \addplot[orange!80!white, fill opacity=\opacity, forget plot, forget plot] fill between[of=GIN-bot and GIN-top];
    \addplot[orange, mark=Mercedes star, solid] table [x=xaxis, y=GIN-A] {\acctable};

    \legend{DCN, DCN-30, DCN-20, DCN-10, GraphSAGE, GIN}
\end{axis}
\end{tikzpicture}
	\end{subfigure}
	\begin{subfigure}{0.32\textwidth}
		\centering
		\begin{tikzpicture}[baseline,scale=1]

\pgfplotstableread{data/density/density_inf-random-med_acc.csv}\acctable
\pgfplotstableread{data/density/density_inf-random-prct70_acc.csv}\prcttop
\pgfplotstableread{data/density/density_inf-random-prct30_acc.csv}\prctbot

\pgfmathsetmacro{\opacity}{0.2}
\pgfmathsetmacro{\contourop}{0.25}

\begin{axis}[
    xlabel={(c) Edge Probability},
    xmin=.1,
    xmax=.7,
    xtick = {.1, .3, ..., .7},
    ylabel={Accuracy},
    ymin = 0,
    ymax = 1.03,
    ytick = {0, .2, ..., 1},
    grid style=densely dashed,
    grid=both,
    legend style={
        at={(.5, 1.02)},
        anchor=south},
    legend columns=3,
    width=175,
    height=140,
    ]

    \addplot [blue!90!black, name path = DCN-bot, opacity=\contourop, forget plot] table [x=xaxis, y=DCN] \prctbot;
    \addplot [blue!90!black, name path = DCN-top, opacity=\contourop, forget plot] table [x=xaxis, y=DCN] \prcttop;
    \addplot[blue!90!black, fill opacity=\opacity, forget plot] fill between[of=DCN-bot and DCN-top];
    \addplot[blue!80!black, mark=o, solid] table [x=xaxis, y=DCN] {\acctable};

    \addplot [red!90!white, name path = GCNN-5-bot, opacity=\contourop, forget plot] table [x=xaxis, y=FB-GCNN-5] \prctbot;
    \addplot [red!90!white, name path = GCNN-5-top, opacity=\contourop, forget plot] table [x=xaxis, y=FB-GCNN-5] \prcttop;
    \addplot[red!90!white, fill opacity=\opacity, forget plot, forget plot] fill between[of=GCNN-5-bot and GCNN-5-top];
    \addplot[red, mark=+, solid] table [x=xaxis, y=FB-GCNN-5] {\acctable};

    \addplot [green!70!black,name path = GraphSAGE-bot, opacity=\contourop, forget plot] table [x=xaxis, y=GraphSAGE-A] \prctbot;
    \addplot [green!70!black,name path = GraphSAGE-top, opacity=\contourop, forget plot] table [x=xaxis, y=GraphSAGE-A] \prcttop;
    \addplot[green!70!black, fill opacity=\opacity, forget plot, forget plot] fill between[of=GraphSAGE-bot and GraphSAGE-top];
    \addplot[green!65!black, mark=x, solid] table [x=xaxis, y=GraphSAGE-A] {\acctable};

    \addplot [blue!60!white, name path = DCN-20-bot, opacity=\contourop, forget plot] table [x=xaxis, y=DCN-20] \prctbot;
    \addplot [blue!60!white, name path = DCN-20-top, opacity=\contourop, forget plot] table [x=xaxis, y=DCN-20] \prcttop;
    \addplot[blue!60!white, fill opacity=\opacity, forget plot] fill between[of=DCN-20-bot and DCN-20-top];
    \addplot[blue!70!white, mark=o, solid] table [x=xaxis, y=DCN-20] {\acctable};

    \addplot [red!60!white, name path = GCNN-2-bot, opacity=\contourop, forget plot] table [x=xaxis, y=FB-GCNN-2] \prctbot;
    \addplot [red!60!white, name path = GCNN-2-top, opacity=\contourop, forget plot] table [x=xaxis, y=FB-GCNN-2] \prcttop;
    \addplot[red!60!white, fill opacity=\opacity, forget plot, forget plot] fill between[of=GCNN-2-bot and GCNN-2-top];
    \addplot[red!70!white, mark=+, dotted] table [x=xaxis, y=FB-GCNN-2] {\acctable};
    
    \addplot [orange!80!black, name path = GIN-bot, opacity=\contourop, forget plot] table [x=xaxis, y=GIN-A] \prctbot;
    \addplot [orange!80!black, name path = GIN-top, opacity=\contourop, forget plot] table [x=xaxis, y=GIN-A] \prcttop;
    \addplot[orange!80!white, fill opacity=\opacity, forget plot, forget plot] fill between[of=GIN-bot and GIN-top];
    \addplot[orange, mark=Mercedes star, solid] table [x=xaxis, y=GIN-A] {\acctable};

    \legend{DCN, FBGNN-5, GraphSAGE, DCN-20, FBGNN-2, GIN}
\end{axis}
\end{tikzpicture}
	\end{subfigure}
    \vspace{-3mm}
	\caption{(a) reports the MNSE in the network diffusion task as the noise in the observations increases; For the source identification task, (b) and (c) depict the influence of increasing respectively the proportion of unobserved nodes and the edge probability.
    We report the median performance across 25 realizations and values between the first and third quartile in the shaded area.}\label{fig:experiments}
    \vspace{-3mm}
\end{figure*}

\vspace{2mm}
\noindent
\textbf{Experiment setup.}
The numerical evaluation is conducted over synthetic data. 
Unless specified otherwise, graphs are sampled from an Erd\H{o}s-Rényi random DAG model with $N=100$ nodes and an edge probability $p=0.2$. The input-output signals are related as $\bby = \bbH\bbx$.
Here, $\bbx$ is a sparse input signal whose non-zero entries are restricted to the first nodes, $\bbH$ is a DAG filter as in \eqref{eq:dag_filter} composed of 25 causal GSOs selected uniformly at random, and $\bby$ is the output of the diffusion of $\bbx$ across the DAG.
We create $M=2000$ input-output pairs of DAG signals and split them as 70\% for training, 20\% for validation, and 10\% for testing.

We consider two learning problems.
First, we address the task of learning a \emph{network diffusion} process where, given a new sparse input $\bbx_{test}$, we aim at estimating the associated output $\bby_{test}$.
The observed input and output signals are corrupted with zero-mean additive white Gaussian noise with a normalized power of $0.05$.
This task amounts to solving a regression problem where we consider the MSE as the loss function $\ccalL$, and report the normalized MSE (NMSE), computed as $\frac{1}{T}\sum_{t=1}^T\| \bby_t - \hby_t \|_2^2 / \| \bby_t \|_2^2$ over $T$ test signals.
The second task tackles the \emph{source identification} problem, where we are given a partially observed output signal $\bby_{test}$ and the goal is to predict which node was the source originating the observation.
Now, the input $\bbx$ contains a single non-zero entry constrained to be located within the first 20 nodes.
Since the diffusion $\bby = \bbH\bbx$ does not alter the signal value at the source nodes, we mask the output $\bby$ so observations from possibly seeding nodes are not available.
The task is cast as a classification problem where the graph label is the index of the source node.
We adopt the cross-entropy loss and report the mean accuracy. 
The reported results comprise 25 independent realizations.

\vspace{2mm}
\noindent
\textbf{Baselines.}
The performance of the DCN is compared with several baselines.
The least-squares (LS) model captures the true linear model, computes an LS  estimate of $\bbh$ from the training data and then predicts test outputs.
As non-linear baselines, we include FB-CGNN, the convolutional GNN based on graph filters from~\cite{ruiz2021graph} [cf. \eqref{eq:fb_gnn}], GCN~\cite{kipf2016semi}, GAT~\cite{velivckovic2018graph}, GIN~\cite{xu2018powerful}, GraphSAGE~\cite{hamilton2017inductive}, and the MLP.
We do not include D-VAE~\cite{zhang2019advances} and DAGNN~\cite{thost2021directed} here (although they are GNNs for DAG-based data), because their training time is prohibitive for moderately large graphs (as here where $N=100$).

\vspace{2mm}
\noindent
\textbf{Preliminary results.}
We start by assessing the overall performance of the models considered in the network diffusion and the source identification problems.
The results collected in \cref{tab:summary_results} showcase the superior performance of the proposed DCN in both tasks.
In the network diffusion setting, we observe how using a subset of 30 or 10 GSOs randomly selected (DCN-30 and DCN-10) decouples the DCN from the number of nodes without significantly hindering the performance; DCN-30 is the second-best alternative.
Comparing DCN and LS, the superior performance of DCN is due to a major resilience to the noise in the inputs, as will be evident in the upcoming experiments.
As expected, when using the transposed matrices $\bbT_k^\top$ in DCN-T (i.e., when following the reverse ordering of the DAG), the performance drops significantly, especially when using only a subset of the GSOs.
In contrast, moving on to the source identification task we observe that while the DCN variants are incapable of identifying the source node, the alternatives DCN-T have an almost perfect accuracy.
Intuitively, this reflects how for identifying the source nodes from the output one needs to navigate the DAGs in the reverse order as facilitated by the transposed GSOs.
In conclusion, this highlights the importance of properly harnessing the directionality of the DAG.
Note that for the baselines we use the GSO with the best performance.

\vspace{2mm}
\noindent
\textbf{Test case 1.}
The results depicted in \cref{fig:experiments}a reflect how increasing the power of the noise in the observed signals impacts the performance of the network diffusion problem.
In the absence of noise, the LS model outperforms the alternatives, but its performance deteriorates swiftly as the noise power increases.
In contrast, the proposed DCN attains a similar NMSE to that of the LS in the noiseless setting and its performance is more robust and in the presence of noise.
This experiment showcases that i) the proposed DCN obtains a performance comparable to that of the optimal solution (LS); and ii) assuming a non-linear model even though the true generative model is linear provides a solution more robust to noise.

\vspace{2mm}
\noindent
\textbf{Test Case 2.}
Moving on to the source identification problem, \cref{fig:experiments}b evaluates the accuracy of the DCN when a different number of GSOs is employed as the proportion of unobserved nodes increases.
Based on the results of \cref{tab:summary_results}, all the methods use the transposed of their respective GSO.
It is worth mentioning that higher values on the x-axis render a more challenging setting since: i) fewer nodal observations are available; and ii) every unobserved node is allowed to be a source node, hence increasing the complexity of the classification problem.
First, we note that the accuracy of DCN remains high even for large fractions of unobserved nodes, demonstrating its robustness.
From the performance of DCN-30, DCN-20, and DCN-10, the trade-off between reducing the number of GSOs and maintaining a high accuracy and low variability is evident.
Nonetheless, DCN-30 shows a similar performance to that of DCN up to observing 50\% of the nodes, and DCN-10 outperforms the non-DAG-based alternatives. 

\vspace{2mm}
\noindent
\textbf{Test Case 3.}
Finally, we evaluate the impact of the density of the graph on the source identification problem.
To that end, we measure the accuracy as the edge probability increases (\cref{fig:experiments}c), where we observe that a denser graph poses a greater challenge to identify the source.
Intuitively, as the graph gets closer to a fully connected DAG, the signal generated from different nodes becomes more homogeneous, with the discrepancies being due only to the distance between seeding nodes.
In addition, this experiment shows that both DCN and DCN-20 consistently outperform FBGNN using filters of order 2 and 5, emphasizing the benefits stemming from replacing GSP filters with DAG filters.

\section{Conclusions}
In this work, we proposed DCN, a GCNN tailored to the particular structure encoded by DAGs.
At its core lies a novel convolutional layer, which is built upon recent results to process signals defined over DAGs and posets, has ties to linear SEMs, is permutation equivariant, endows the architecture with a spectral representation, and exploits the partial order imposed by the DAG.
Moreover, our preliminary numerical evaluation demonstrated a promising performance and showed that the convolution operation can be detached from the number of nodes, thus addressing a potential limitation.

\bibliographystyle{IEEEbib}
\bibliography{myIEEEabrv,biblio}

\end{document}